\documentclass{article}

\PassOptionsToPackage{numbers, compress}{natbib} 
\usepackage[final]{nips_2017}
\usepackage{nips_2017}
\usepackage[utf8]{inputenc} 
\usepackage[T1]{fontenc}    
\usepackage{hyperref}       
\usepackage{url}            
\usepackage{booktabs}       
\usepackage{amsfonts}       
\usepackage{nicefrac}       
\usepackage{microtype}      

\usepackage{graphicx}
\usepackage{multirow}
\usepackage{tabularx}
\usepackage{makecell}
\usepackage{subfig}

\usepackage{color}
\usepackage{textcomp}

\usepackage{footnote}
\makesavenoteenv{table}

\title{Highrisk Prediction from Electronic Medical Records via Deep Attention Networks}

\author{
  You Jin Kim\textsuperscript{1}, Yun-Geun Lee\textsuperscript{1}, Jeong Whun Kim\textsuperscript{2}, Jin Joo Park\textsuperscript{2}, Borim Ryu\textsuperscript{2}, Jung-Woo Ha\textsuperscript{1} \\
  \textsuperscript{1}Clova AI Research, NAVER Corp., Seongnam, Korea \\
  \textsuperscript{2}Seoul National University Bundang Hospital, Seongnam, Korea \\
}

\begin{document}

\maketitle

\begin{abstract}
Predicting highrisk vascular diseases is a significant issue in the medical domain. Most predicting methods predict the prognosis of patients from pathological and radiological measurements, which are expensive and require much time to be analyzed. Here we propose deep attention models that predict the onset of the high risky vascular disease from symbolic medical histories sequence of hypertension patients such as ICD-10 and pharmacy codes only, Medical History-based Prediction using Attention Network (MeHPAN). We demonstrate two types of attention models based on 1) bidirectional gated recurrent unit (R-MeHPAN) and 2) 1D convolutional multilayer model (C-MeHPAN). Two MeHPAN models are evaluated on approximately 50,000 hypertension patients with respect to precision, recall, f1-measure and area under the curve (AUC). Experimental results show that our MeHPAN methods outperform standard classification models. Comparing two MeHPANs, R-MeHPAN provides more better discriminative capability with respect to all metrics while C-MeHPAN presents much shorter training time with competitive accuracy. 


\end{abstract}

\section{Introduction}

Medical histories of patients such as diagnosis, prescription, and medication records are important data when doctors diagnose diseases~\cite{Summerton273}. These records characterize the current status and the disease progress of patients. 
Most methods for diseases diagnosis have mainly focused on pathological and radiological measurements as well as clinical information for predicting the current status and prognosis of the patients~\cite{lung_cnn, thorax_cnn}. However, the measurement-based methods usually require considerable costs to be analyzed with respect to the time and the price. 


The deep learning~\cite{lecun2015deep} has been successfully applied to various fields such as speech signal processing~\cite{arik2017deep, amodei2016deep}, natural language processing~\cite{lin2017structured, gehring2017convolutional} and vision processing~\cite{huang2016densely, redmon2016yolo9000}. Recently, successful studies using deep learning are reported in biomedical domains including disease prediction~\cite{NIPS2016_6321} and gene sequencing~\cite{umarov2017recognition}.

Here we predict the onset of highrisk vascular diseases for hypertension patients, using deep learning algorithms with an attention mechanism~\cite{xu2015show} to analyze patients' medical history, Medical History-based Prediction using Attention Network (MeHPAN). We define cardiovascular and cerebrovascular diseases as the highrisk case, which are significant factors to cause the mortality in the world~\cite{lozano2012global}.
The medical history consists of the diagnosis and the medication sequences that are decided by a doctor. We describe two types of deep attention models using bidirectional gated recurrent unit (GRU)~\cite{chung2014empirical} (R-MeHPAN) and 1D convolutional model~\cite{kim2014convolutional} (C-MeHPAN). The structures enable the model to explicitly learn features from sequential information on the progress of the disease status~\cite{gehring2017convolutional, sutskever2014sequence}, thus predicting the highrisk prognosis from given medical records.
The attention allows the model to focus on the important diagnosis and medication records in the input sequences for predicting the highrisk cases, to train the models more accurately. 
In addition, our models use the multi-task learning by utilizing two separate output layers. 
It is conducted together that the binary classification that predicts whether a patient has a vascular disease and the multi-class classification that classifies the cardiovascular disease, the cerebrovascular disease and no vascular disease.   

We evaluate the performance of two MeHPAN models, measuring running time, precision, recall, f1-measure and area under the curve (AUC). The MeHPANs show better performance than the standard machine learning techniques that are the support vector machine (SVM) and the random forest. R-MeHPAN shows higher performance than C-MeHPAN in every metrics, except for running time. C-MeHPAN is 10 times faster than R-MeHPAN. 

\section{Data Description}
\label{Data}

We use electronic medical data to predict the onset of the vascular disease obtained from Seoul National University Bundang Hospital. The dataset consists of 62,878 patients' medical history. 50,720 patients do not have vascular diseases and 12,158 patients have. Among 12,158 patients, 5,704 have the cardiovascular disease and 6,454 have the cerebrovascular disease~\cite{ha2017predicting}.      
 
A patient's medical history consists of two parts, the diagnosis history and the medication history. 
The diagnosis history consists of the code sequence, the date sequence and the kind sequence. 
The code sequence consists of the codes that correspond to doctors' diagnosis results. The code have 6,667 types. 
The date sequence is comprised of the date that a patient visit the hospital. Using the date and the below equation, we calculate the duration of the patient's disease, and obtain the duration sequence .
\[ DUR_{i} = Max( log( T_{last} - T_{i} + 1 ), 1 ) \]
In the equation, $T_{last}$ is the last date when the patient visits the hospital and $T_{i}$ is the date when the $i$-th code are diagnosed. The kind sequence represents the visit type, which are outpatient, inpatient or emergency visit types. 

The medication history consists of the medication code sequence, the medication period sequence and the medication type sequence. 
The medication code sequence contains prescribed drug information.
The medication period sequence and the following equation are used to calculate the medication duration sequence. 
\[ MD\_DUR_{i} = log( period_{i} + 1 ) \]
The equation calculates the duration of the $i$-th prescribed medicine. The medication type sequence is comprised of each medication's type, which are pill, injection or sap. 

This study was approved by the Institutional Review Board at Seoul National University Bundang Hospital (IRB approval no. “B-1512/326-102"). SNUBH IRB granted waivers of informed consent since this study involved analyses of retrospective data where all patient information was anonymized and de-identified prior to analysis.

\section{Attention Models for Highrisk Prediction}
\label{Models}

\subsection{R-MeHPAN: RNN-based Attention Model}
The GRU is an advanced recurrent neural networks using a gate approach~\cite{chung2014empirical}, similar to long short term memory (LSTM). GRU does not use an explicit memory cells unlike the LSTM. This relative simple structure allows GRU to be more efficient than the LSTM. 

\begin{figure}[t]
\centering
  \includegraphics[width=0.85\columnwidth]{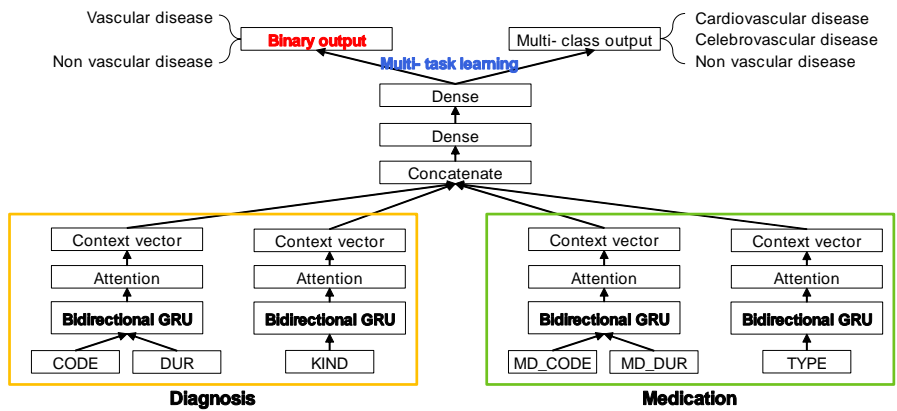}    
  \caption{R-MeHPAN structure.}
  \label{fig::gru}
\end{figure}

We use multiple feature-specific bidirectional GRUs~\cite{ha2016large} with attention for R-MeHPAN that predicts the high-risk of hypertension patients. 
R-MeHPAN consists of two modules such as the diagnosis module and the medication module, which are dedicated to each GRU, as shown in Figure~\ref{fig::gru}. 

\noindent\textbf{Diagnosis module} \textit{CODE} and \textit{DUR} that denote the diagnosis code sequence and the diagnosis duration sequence respectively, are embedded together for their associations, and these embedded vectors are given to the bidirectional GRU as input. 
The attention is introduced to focus on the important diagnosis information for prediction. As shown in Figure~\ref{fig::1-d} (a), we use a two-layer feedforward network for the attention, calculated by element-wise multiplying the embedding vectors and the GRU outputs, thus generating context vectors.
The diagnosis kind sequence, \textit{KIND} is embedded separately, and goes to the different bidirectional GRU. Each \textit{KIND} can be one among three values such as outpatient, inpatient, and emergency. 

\noindent\textbf{Medication module} The structure of the medication part is similar to the diagnosis part, except input. The medication code sequence, \textit{MD\_CODE} and the medication duration sequence, \textit{MD\_DUR} are embedded together. The bidirectional GRU learns the sequential patterns of the medication records. We get the context vector using the same attention mechanism to the diagnosis module. 
The medication type sequence, \textit{TYPE} is embedded separately, and it is used as the input of the GRU. Similarly, we conduct attention and then calculate the context vector. 

Four context vectors generated from the diagnosis the medication modules are concatenated, and then are given to the output layer for calculating the probability of each class through two dense layers.

We apply the multi-task learning to the proposed model, in which the model is trained for related tasks at the same time, giving the model additional information. We conduct the binary classification and the multi-class classification simultaneously. In the binary classification, we predict whether hypertension patients have the highrisk vascular disease or not. In the multi-class classification, we classify three cases, that are the cardiovascular disease, the cerebrovascular disease and no vascular disease. These multi-task learning setup allows our model to use the error information between two vascular diseases. We measure the performance of the binary classification only in this study.

\subsection{C-MeHPAN: 1D Convolution-based Attention Model}
One dimensional (1D) convolution is the convolution calculation with the one dimensional kernel. It can be used for the single dimensional data, such as time series data. The 1D convolution does not depend on the previous time steps’ calculations like GRU. Therefore, it is more efficient in the parallel processing and shows faster speed than the GRU~\cite{kim2014convolutional}.

\begin{figure}[t]
\centering
  \subfloat[Attention mechanism]{\includegraphics[width=0.4\columnwidth]{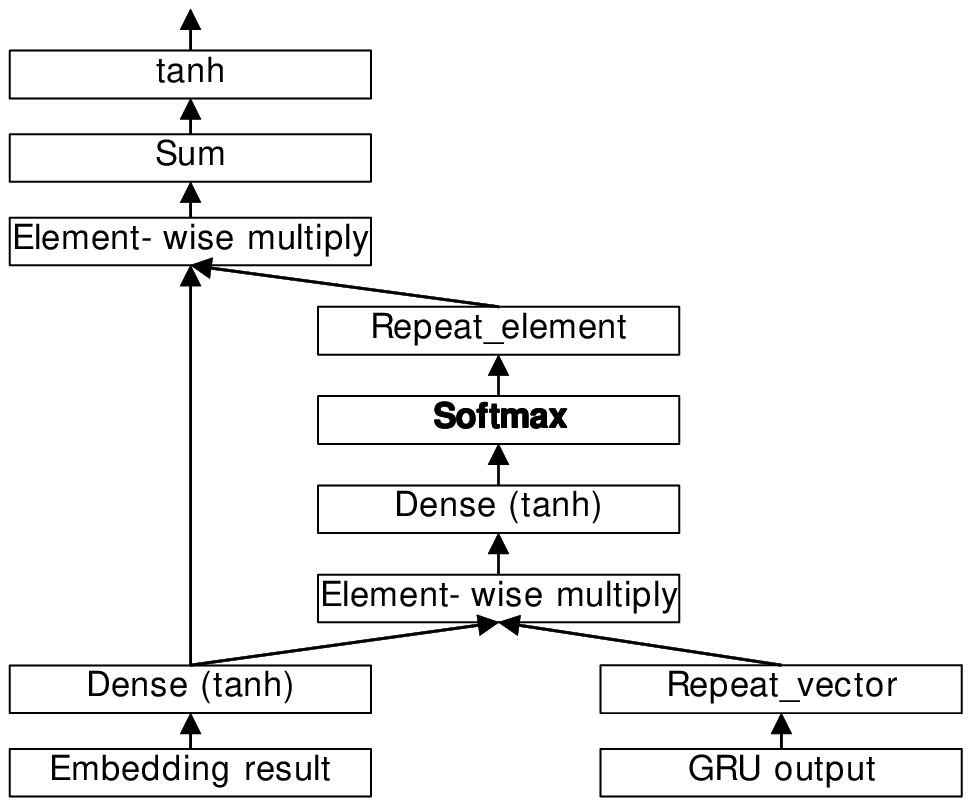}}
  \subfloat[1D convolution~\cite{gehring2017convolutional}]{\includegraphics[width=0.5\columnwidth]{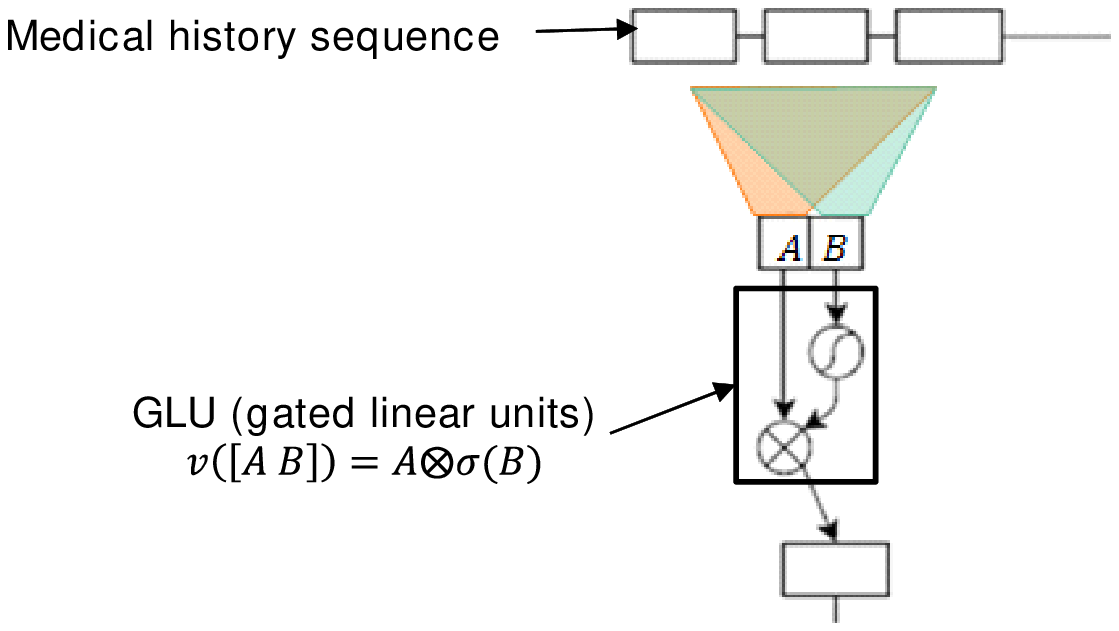}}
  \caption{C-MeHPAN structure.}
  \label{fig::1-d}
\end{figure}

We apply the 1D convolution instead of the bidirectional GRU. Figure~\ref{fig::1-d} illustrates C-MeHPAN structure. We use two 1D convolutions for one time step. One convolution calculates A and the other calculates B for the gated linear units (GLU)~\cite{gehring2017convolutional, glu}. To apply the attention in Figure~\ref{fig::1-d} (a), we convert the dimension of the 1D convolution from 3 to 2 in three ways. 
First, We sum the 3 dimensional output along the time step axis. 
Second, We calculate weighted sum, using the weight vector that has the higher weight for the later time step.  
Third, We use the 1D convolution output’s the last time step value. 
We calculate the attention using one of three methods, and then obtain context vectors. 
We obtain four context vectors, same as the attention mechanism of R-MeHPAN. We concatenate these four context vectors, and perform the multi-task learning like R-MeHPAN.

\section{Experimental Results}
\label{Experiments}
We use four prediction and one time metrics such as precision, recall, F1-measure, area under the curve (AUC) and running time for model evaluations. 
All prediction metrics are focused on highrisk vascular disease class. AUC is measured because of the robustness to the imbalanced data. Training time is measured, considering the practical usage in the real world. All Experiments are conducted on a GPU server with NVIDIA Tesla P40.

Patients' medical history sequence have different length, because the number of patients' visit is diverse. We add zero padding to the medical history sequence to make the length identical. 
We evaluate the models by 10 set separation, and the train data to test data ratio is eight to two.



%

Experimental results can be found in Table~\ref{table::performance}.
SVM and random forest show high precision but they show very low recall. It indicates that they do not predict well the onset of the vascular disease from the medical history. 
Two MeHPAN shows better performance than SVM and random forest. R-MeHPAN shows better performance than C-MeHPAN in every metric except for running time. C-MeHPAN is about 10 times faster than R-MeHPAN, with competitive accuracy.

In Table~\ref{table::performance}, we can also check the experimental results of the three methods how the 3 dimensional output of C-MeHPAN is converted to 2 dimension. Weighted sum method shows slightly better performance than the others in every metrics except for running time. Last time step method shows the best running time but the difference is very small.

\begin{table}
  \caption{Comparison of highrisk prediction for each model}
  \label{table::performance}
  \centering
  \begin{tabularx}{\textwidth}{X|c|cccccc}
  \toprule
  \multicolumn{2}{c|}{} & \makecell{Running time \\ (/min)} & Precision & Recall & F1-measure & AUC \\ \midrule
  \multicolumn{2}{c|}{SVM} & 78.02172 & 1.00000 & 0.02212 & 0.04328 & 0.51106 \\ \midrule
  \multicolumn{2}{c|}{Random forest} & 0.09290 & 0.86850 & 0.43689 & 0.58130 & 0.71051 \\ \midrule
  \multicolumn{2}{c|}{R-MeHPAN} & 65.50967 & 0.71831 & 0.77549 & 0.74567 & 0.85123 \\ \midrule
  \multirow{3}{*}{C-MeHPAN} & Sum & 7.01759 & 0.64964 & 0.74154 & 0.69217 & 0.82268 \\ \cline{2-7}
                                          & Weighted sum & 7.07990 & 0.66065 & 0.74828 & 0.70164 & 0.82807 \\ \cline{2-7}
                                          & Last time step & 6.97894 & 0.65180 & 0.74842 & 0.69660 & 0.82625 \\ 
  \bottomrule
  \end{tabularx}
\end{table}

\section{Conclusions and Future Works}
\label{Conclusions}
Here we proposed R-MeHPAN and C-MeHPAN to predict the onset of highrisk vascular diseases from the medical records of hypertension patients. Two MeHPANs showed competitive performance by analyzing the medical history data only. R-MeHPAN shows higher accuracy than C-MeHPAN whereas C-MeHPAN spends much less time on training than R-MeHPAN.

We will improve the way that handles the variable length medical history. We will investigate how the attention has influence on model performance and interpret its score for explainability. Additionally, we have a plan to predict when the vascular disease occurs, in order to alert hypertension patients.


\subsubsection*{Acknowledgments}
This work was supported by the Creative Industrial Technology Development Program (10053249) funded by the Ministry of Trade, Industry and Energy (MOTIE, Korea).

\small
\bibliographystyle{IEEEbib.bst}
\bibliography{nips}

\end{document}